\documentclass[conference]{IEEEtran}
\IEEEoverridecommandlockouts
\usepackage{cite}
\usepackage[hidelinks]{hyperref}
\usepackage{amsmath,amssymb,amsfonts}
\usepackage{algorithmic}
\usepackage{graphicx}
\usepackage{textcomp}
\usepackage{xcolor}
\usepackage{subcaption}
\usepackage{float}
\usepackage{tabularx}
\usepackage{booktabs}
\usepackage{url} 
\usepackage{multirow}
\usepackage[dvipsnames]{xcolor}
\usepackage[table]{xcolor}

\def\BibTeX{{\rm B\kern-.05em{\sc i\kern-.025em b}\kern-.08em
    T\kern-.1667em\lower.7ex\hbox{E}\kern-.125emX}}
\begin{document}

\title{{\footnotesize \textit{To appear in Proceedings of the 2025 IEEE International Conference on Big Data}\\[1.0em]}\textsc{DeepSalt}: Bridging Laboratory and Satellite Spectra through Domain Adaptation and Knowledge Distillation  for Large-Scale Soil Salinity Estimation

\thanks{This research was supported by the National Science Foundation (1931363, 2312319), the National Institute of Food Agriculture
(COL014021223), an NSF/NIFA Artificial Intelligence Institutes AI-LEAF [2023-03616].}
}

\author{\IEEEauthorblockN{Rupasree Dey\textsuperscript{1}, Abdul Matin\textsuperscript{1}, Everett Lewark\textsuperscript{1}, Tanjim Bin Faruk\textsuperscript{1}, Andrei Bachinin\textsuperscript{1},\\ Sam Leuthold\textsuperscript{2}, M. Francesca Cotrufo\textsuperscript{2}, Shrideep Pallickara\textsuperscript{1}, and Sangmi Lee Pallickara\textsuperscript{1}}
\IEEEauthorblockA{\textsuperscript{1}Department of Computer Science, Colorado State University, USA\\
\textsuperscript{2}Department of Soil and Crop Sciences, Colorado State University, USA\\
\{rupasree.dey, abdul.matin, elewark, tanjim.faruk, andrei.bachinin, \\
sam.leuthold, francesca.cotrufo, shrideep.pallickara, sangmi.pallickara\}@colostate.edu}
}

\maketitle

\thispagestyle{plain}
\pagestyle{plain}

\begin{abstract}
Soil salinization poses a significant threat to both ecosystems and agriculture because it limits plants' ability to absorb water and, in doing so, reduces crop productivity. This phenomenon alters the soil's spectral properties, creating a measurable relationship between salinity and light reflectance that enables remote monitoring. While laboratory spectroscopy provides precise measurements, its reliance on in-situ sampling limits scalability to regional or global levels. Conversely, hyperspectral satellite imagery enables wide-area observation but lacks the fine-grained interpretability of laboratory instruments. To bridge this gap, we introduce \textsc{DeepSalt}, a deep-learning-based spectral transfer framework that leverages knowledge distillation and a novel \textit{Spectral Adaptation Unit} to transfer high-resolution spectral insights from laboratory-based spectroscopy to satellite-based hyperspectral sensing. Our approach eliminates the need for extensive ground sampling while enabling accurate, large-scale salinity estimation, as demonstrated through comprehensive empirical benchmarks.  \textsc{DeepSalt} achieves significant performance gains over methods without explicit domain adaptation, underscoring the impact of the proposed \textit{Spectral Adaptation Unit} and the knowledge distillation strategy. The model also effectively generalized to unseen geographic regions, explaining a substantial portion of the salinity variance.
\end{abstract}


\begin{center}

© 2025 IEEE. Personal use of this material is permitted. Permission from IEEE must be obtained for all other uses, in any current or future media, including reprinting/republishing this material for advertising or promotional purposes, creating new collective works, for resale or redistribution to servers or lists, or reuse of any copyrighted component of this work in other works.

\end{center}

\section{Introduction}

Soil salinization poses a significant threat to agriculture and ecosystems. The accumulation of salt creates osmotic stress \cite{b1}, which reduces the uptake of plant water and nutrients, reduces crop yields, and accelerates land degradation. These impacts are most pronounced in arid and semi-arid regions, where salinity hotspots directly undermine food security.

Remote sensing offers an efficient pathway for monitoring salinity at large scales. Compared with costly and sparse field sampling, satellite observations provide wall-to-wall coverage and a measurable link between spectral reflectance and soil salinity. Among these, hyperspectral data \cite{b7} have emerged as a powerful tool to map salinity dynamics due to their ability to capture subtle spectral features. However, the salinity estimation poses unique challenges. Unlike vegetation (with strong red-edge signals) or water (with distinct absorption features), salts such as NaCl produce weak responses in satellite-accessible ranges \cite{sahbeni}. Salinity distributions obtained from remote sensing observations exhibit high skewness: Most pixels show negligible salinity ($<1$ dS/m), while critical hotspots ($>10$ dS/m) are rare but essential to detect. These properties create significant challenges for model generalization. The difficulty lies in extracting faint, nonlinear signals from hyperspectral imagery under conditions of extreme class imbalance.


Hyperspectral satellite imaging captures data across a large number of narrow, contiguous spectral bands. It encompasses a broad spectral range: from 420 to 1,000 nm in the visible to near-infrared (VNIR), and from 900 to 2,450 nm in the shortwave infrared (SWIR). This high radiometric resolution and stability across these extensive spectral regions enable a detailed analysis of material composition and surface properties.

Although hyperspectral satellite imagery offers the potential for high-throughput soil analysis, its effectiveness is often limited due to vegetation, coarse debris, or other objects covering the soil surface. These surface obstructions interfere with reflectance signals, making it difficult to isolate the spectral signatures of bare soil. 

Supervised learning depends on accurate ground truth, which in soil salinity estimation usually means laboratory measurement of electrical conductivity ($EC_e$) from a saturated paste extract. This reliance on lab data creates a domain adaptation challenge: while satellite spectra are noisy and environmentally perturbed, laboratory spectra are clean and collected under carefully controlled conditions. The gap between them is further widened by spectral coverage---Laboratory FTIR spectroscopy operates in the mid-infrared (MIR: 2,500--25,000 nm) and captures molecular signatures of salts, absent from the VNIR–SWIR range accessible to satellites. Ancillary data such as soil texture, climate, and topography introduce another layer of complexity. Although they provide complementary insights into soil processes, their varied scales and modalities require deliberate integration to avoid bias and ensure effective learning. Together, these challenges point toward the need for new machine learning approaches that can transfer knowledge across sensing modalities while incorporating heterogeneous data sources. This fusion challenge reflects the broader difficulty of integrating multimodal, multi-scale data into coherent models that can yield actionable insights.

Previous efforts have explored statistical and machine learning models (Random Forests, Partial Least Squares regression, Support Vector Regression, and several deep learning approaches) to estimate salinity from remote sensing data. Although traditional methods such as Random Forest and PLS have offered interpretability and robustness, they falter in capturing the nonlinear complexities characteristic of hyperspectral data, underscoring the need for more flexible architectures. There is a need for models that not only learn from large-scale high-dimensional data but also transfer domain-specific knowledge across disparate modalities: spectroscopy, satellite reflectance, and ancillary environmental variables.

\textbf{Research Questions}
that we explore include:

\textbf{RQ-1}: How can hyperspectral satellite imagery be effectively utilized to estimate soil salinity at scale?

\textbf{RQ-2}: What approaches enable a robust transfer of knowledge from precise laboratory salinity measurements to models trained on noisier, satellite-derived observations? 

\textbf{RQ-3}: How can transformer-based deep learning models capture the subtle spectral signatures of soil salinity across varied environmental and geographic settings?

\subsection{Approach Summary}
We propose a novel deep learning framework to estimate the salinity level of the soil using hyperspectral satellite imagery based on a Transformer architecture. The core challenge lies in reliably extracting soil salinity signals from noisy spectral reflectance data, which can be distorted by surface conditions and environmental factors. To address this, our model (referred to as \textsc{DeepSalt}) is designed to learn spectral characteristics from a model trained on laboratory-based spectral absorbance data. Reflectance and absorbance account for the total interaction of light with a material: reflectance refers to the fraction of light reflected from a surface, while absorbance represents the amount of light absorbed by the material. Together with transmittance, these values should sum up to 100\%. We posit that by learning from these well-characterized absorbance signatures, the model can more effectively recognize similar patterns in reflectance data gathered via satellite, despite the presence of noise. This will allow for a more accurate and scalable estimate of soil salinity.

To enable accurate soil salinity estimation, \textsc{DeepSalt} incorporates a tailored knowledge distillation strategy. Knowledge distillation typically involves transferring knowledge from a large, high-performing model (the ``teacher") to a smaller, more efficient model (the ``student"). The \textsc{DeepSalt} framework adapts this concept in a novel way: rather than focusing on model size, we transfer knowledge from a high-accuracy single-modal model to a multi-modal model trained on noisier, more variable data.  
The teacher model is trained using high-quality spectral data and associated soil salinity values gathered in laboratory settings, where spectroscopy and chemical analyses provide precise measurements. The teacher model captures robust relationships between spectral characteristics and salinity levels. 
The multimodal student model is trained on satellite-based reflectance data, which is inherently noisier and subject to environmental variability. Through the distillation process, the student model assimilates spectral insights learned by the teacher, improving its ability to estimate soil salinity from complex hyperspectral satellite imagery. 

One of the key challenges in our approach is that satellite spectral measurements and laboratory spectroscopy differ fundamentally in their spectral resolution, range, and measurement style. To address this, \textsc{DeepSalt} provides a unique spectral domain adaptation strategy, \textit{Spectral Adaptation Unit} (SAU), which generates latent space to enable cross-sensor harmonization. The SAU is an encoder trained on both spectroscopy and satellite spectral measurements. It is initially trained using high-quality spectroscopy data and subsequently calibrated using paired spectral measurements collected from geospatially proximate locations. This process enables the encoder to align and adapt spectral representations across the two domains.

\noindent\textbf{Paper Contributions}
The contributions of this work include:
\begin{itemize}
    \item We propose \textsc{DeepSalt}, a transformer-based deep neural network specifically designed to capture subtle soil salinity signatures from high-dimensional hyperspectral satellite imagery.
    
    \item We introduce a spectral knowledge distillation framework that transfers domain insights from controlled laboratory absorbance data to satellite reflectance training, enabling robust signal extraction despite real-world noise.
    
    \item We perform extensive empirical evaluations, comparing \textsc{DeepSalt} against traditional machine learning methods and deep learning architectures. Across these benchmarks, \textsc{DeepSalt} consistently achieves superior predictive accuracy.
    
\end{itemize}

\noindent\textbf{Translational Impacts} of this work include: \textbf{(1)}
By bridging controlled laboratory measurements with satellite-based sensing, our methodology offers a generalizable framework for cross-domain modeling. This opens new avenues for other environmental applications (e.g., carbon and moisture estimation) where field data are sparse or noisy.
\textbf{(2)} Because soil salinity signals broader patterns of land degradation under climate stress, our framework can serve as a foundation for early-warning systems, offering timely, spatially resolved soil health insights.
\textbf{(3)} In agricultural contexts, \textsc{DeepSalt} enables scalable, high-resolution salinity mapping, equipping farmers and land managers with actionable data to optimize irrigation, improve crop resilience, and promote sustainable land use.

\section{\textbf{Related Work}}
\label{related-work}

\subsubsection{Machine Learning for Spectral Analysis in Digital Soil Mapping} Digital Soil Mapping (DSM) produces spatially explicit, digitally referenced soil datasets by combining field and laboratory soil observations with environmental information through modeling approaches. Recently, DSM has increasingly adopted machine learning techniques for spectral analysis. The approaches include tree-based ensembles (e.g., Random Forests \cite{Chuanmei}, XGBoost \cite{bioclimatic}), kernel-based methods (e.g., Support Vector Regression \cite{support}), and deep learning architectures (notably CNNs \cite{Mohammadifar} and LSTMs \cite{hochreiter1997long}). These machine learning methods have demonstrated superior performance over traditional approaches.

Among deep learning approaches, Transformer architectures \cite{vaswani2017attention} have rapidly advanced remote sensing spectral analysis \cite{hong2021spectralformer}, offering distinct advantages through their self-attention mechanisms and hybrid designs. Although these methods have shown promising performance in general spectral analysis tasks, transformer applications specifically for soil salinity estimation remain limited compared to their use in predicting other soil properties. Our \textsc{DeepSalt} framework addresses this gap by introducing a novel transformer-based knowledge distillation approach that bridges laboratory-grade FTIR spectroscopy with satellite hyperspectral imaging, a critical innovation for scalable salinity monitoring that simultaneously overcomes the domain shift challenges between controlled experiments and field conditions.

\subsubsection{Multimodal Fusion for Soil Salinity Estimation} Recently, Sui et al. \cite{Sui} demonstrated the effectiveness of multimodal fusion for soil salinity mapping with an $R^2$ value of 0.79 by combining multi-temporal Landsat-8 multispectral satellite data with topographic and groundwater covariates through a CNN. This approach identifies topographic characteristics as the most predictive covariate. Tola et al. \cite{tola} showed that fusing Sentinel-1/2 multispectral satellite data with topographic indices in tree-based models outperforms radar-only approaches for agricultural soils. Our framework improves upon these approaches by: (1) integrating fine-grained soil texture and temporal weather covariates absent in previous work, and (2) introducing transformer-based knowledge distillation to bridge the critical lab-hyperspectral satellite domain gap---addressing a fundamental limitation where existing methods either assume aligned spectral ranges \cite{Sui} or focus solely on field-scale covariates \cite{tola}. This enables robust cross-domain salinity estimation where traditional fusion methods fail.

\subsubsection{Comparative Analysis of Cross-Domain Adaptation Methods} Unlike prior cross-domain adaptation approaches for categorical tasks like crop mapping \cite{Mohammadi} or land cover classification (ST-DASegNet \cite{disentangled}, FLDA-NET \cite{Meng}), our framework addresses the unique challenges of continuous soil salinity estimation. Although existing methods operate within aligned spectral domains (e.g., VIS-SWIR) using style transfer or mutual information, we bridge non-overlapping sensing regimes---laboratory FTIR (2,500–-25,000 nm) and satellite hyperspectral (420–-2,450 nm)---through a domain-informed Spectral Adaptation Unit and geospatially constrained latent space alignment. Crucially, we integrate ancillary soil and weather data to disentangle salinity signals from environmental confounders, a requirement absent in classification-centric methods. Our knowledge distillation strategy further distinguishes itself by transferring noise-invariant spectral patterns from controlled labratory data to noisy satellite observations via layer-wise transformer feature alignment, enabling quantitative regression of highly skewed salinity values (0--90 dS/m) rather than discrete class prediction.

\begin{figure*}[t]
    \centering
    \begin{subfigure}[b]{0.48\textwidth}
        \centering
        \includegraphics[width=\textwidth]{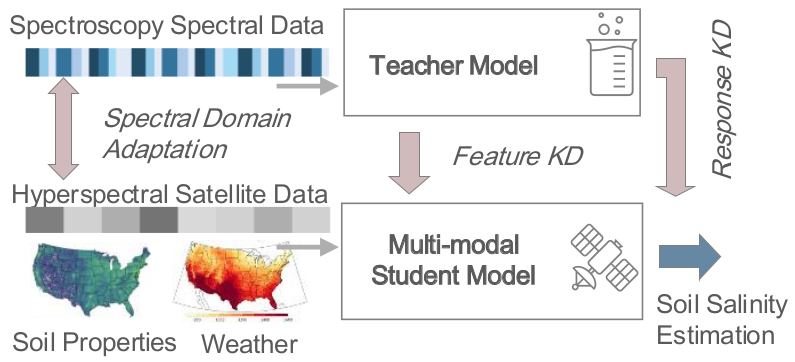}
        \caption{High-level overview of the three-stage \textsc{DeepSalt} pipeline.}
        \label{fig:overview} 
    \end{subfigure}
    \hfill
    \begin{subfigure}[b]{0.48\textwidth}
        \centering
        \includegraphics[width=\textwidth]{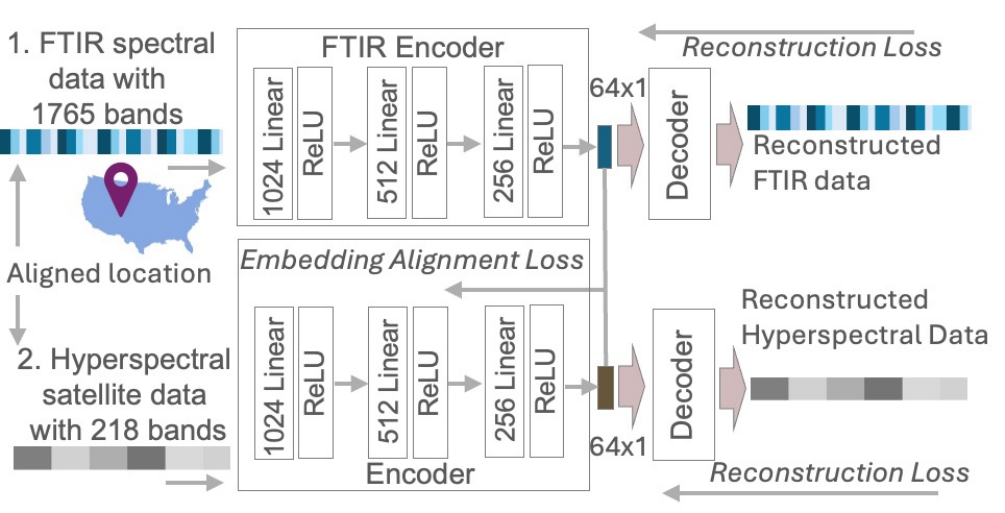}
        \caption{Architecture of the Spectral Adaptation Unit (SAU) for cross-domain latent space alignment (Section \ref{spectral-adaptation}).}
        \label{fig:model-123} 
    \end{subfigure}
    
    \vspace{5mm}

    \begin{subfigure}[b]{\textwidth}
        \centering
        \includegraphics[width=0.8\textwidth]{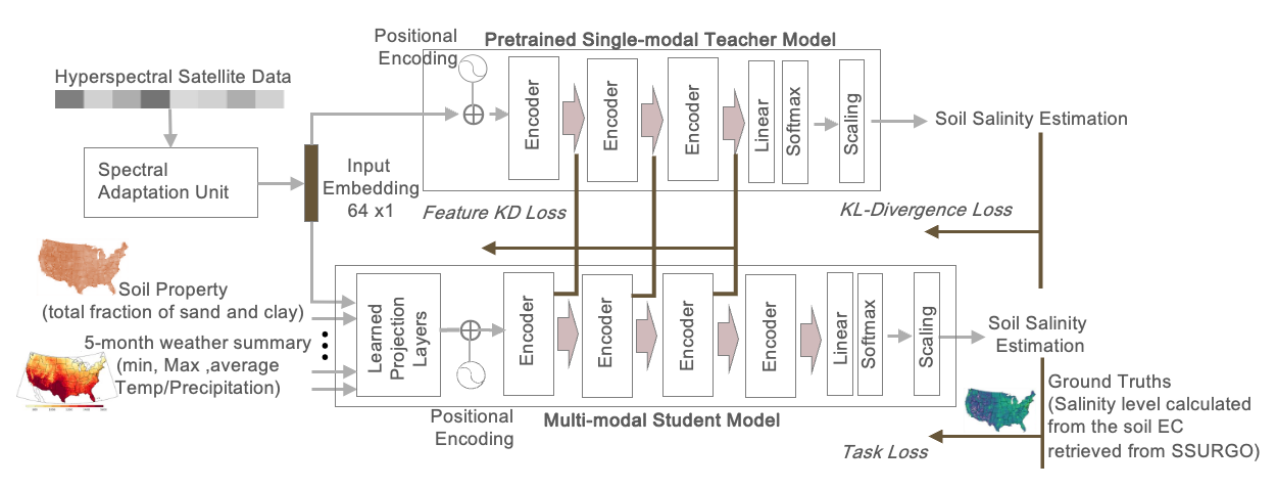}
        \caption{Detailed knowledge distillation process from the teacher to the student model (Section \ref{student-model}).}
        \label{fig:training} 
    \end{subfigure}
    
    \caption{Comprehensive architecture of the \textsc{DeepSalt} framework. (\subref{fig:overview}) illustrates the end-to-end workflow: pretraining a teacher on lab FTIR data, spectral adaptation, and distilling knowledge into a student that fuses adapted satellite spectra with ancillary environmental data. (\subref{fig:model-123}) details the Spectral Adaptation Unit (SAU), which bridges the domain gap between laboratory (FTIR, MIR) and satellite (EnMAP, VNIR-SWIR) spectra by projecting them into a shared latent space. (\subref{fig:training}) shows the multi-objective distillation strategy that transfers the teacher's feature representations and output distributions to the student model.}
    \label{fig:full_architecture} 
\end{figure*}

\section{\textbf{Methodology}}
\label{methodology}

To address the challenge of estimating soil salinity at scale, we propose a deep learning framework, \textsc{DeepSalt}, which relies on hyperspectral satellite imagery and a Transformer-based architecture. A key challenge in this setting lies in distinguishing subtle context-dependent salinity fingerprints from noise caused by environmental factors such as vegetation, soil moisture, and surface roughness that degrade spectral reflectance signals.

Our methodology seeks to overcome this challenge by transferring knowledge from a controlled to a more variable domain. We first train a teacher model on laboratory spectroscopy data, where minimized external interference allows it to capture core spectral relationships between absorbance patterns and salinity levels. Section \ref{pretrain-teacher} details the construction of this model and the extraction of latent representations that encode spectral signatures relevant to salinity estimation.
However, laboratory and satellite spectral measurements differ significantly in format and domain. To bridge this gap, we introduce a spectral domain adaptation technique, detailed in Section \ref{spectral-adaptation}. This adaptation component ensures that the spectral features extracted from satellite imagery can be meaningfully aligned with the representations learned from laboratory data, thereby making knowledge transfer feasible.
Finally, in Section \ref{student-model}, we bring these components together through a tailored knowledge distillation framework. Here, the insights captured by the teacher model are transferred to a multimodal student model that operates on satellite imagery and ancillary data, such as soil properties. This student model learns to estimate soil salinity even in the presence of noisy input, guided by the knowledge distilled from its teacher. Through this three-stage process: representation learning from laboratory data, spectral domain alignment, and knowledge-guided multimodal training, we build a robust and scalable framework for estimating soil salinity in real-world settings.

An overview of our methodology is depicted in Figure \ref{fig:overview}. \textsc{DeepSalt} draws on the spectral insights captured in a noise-free laboratory setting through high-resolution spectroscopy. The model learns the underlying relationships between soil salinity and the spectral characteristics of soil samples---relationships that are often difficult to isolate in field conditions due to confounding environmental noise. These learned associations form the foundation of our knowledge distillation process.
Through this knowledge distillation process, we transfer what the teacher model has learned to a multimodal student model, enabling it to make salinity predictions from hyperspectral satellite imagery and ancillary information. Both teacher and student models use the Transformer architecture \cite{vaswani2017attention}, which excels at processing sequence data by simultaneously modeling interrelationships via self-attention. The Transformer operates through self-attention mechanisms that weigh the importance of each input position relative to others, enabling parallel processing and superior modeling of long-range dependencies, making it especially powerful for capturing complex spectral correlations across wavelengths. This allows the models to efficiently capture both local and global dependencies across spectral bands. The Transformer's multi-head attention and positional encoding are particularly well-suited for spectral analysis, enabling the modeling of diverse patterns while preserving crucial wavelength-order information.




\subsection{Teacher Model Pretraining [RQ-1]}
\label{pretrain-teacher}

The first stage of \textsc{DeepSalt} establishes a strong spectral foundation by training a teacher model on high-quality FTIR spectroscopy data. These laboratory spectra contain 1,765 bands, providing rich detail on salinity-relevant chemical and mineralogical patterns. To reduce dimensionality while retaining diagnostic features, the spectra are first compressed into a 64-dimensional latent representation by the Spectral Adaptation Unit (SAU). This embedding not only facilitates efficient learning but also provides a common basis for later alignment with satellite spectra.

On top of these embeddings, the teacher employs a Transformer encoder with 3 layers, 4 attention heads, and 128-dimensional feedforward networks. Sinusoidal positional encodings ensure that wavelength ordering is preserved, allowing the model to capture both localized absorption features and long-range dependencies across the spectral sequence. Predictions are produced by averaging sequence outputs and passing them through a linear layer with scaled sigmoid activation. The scaling ensures that outputs remain within realistically valid salinity levels (0.05--90 dS/m), thereby producing realistic yet differentiable predictions.

\subsection{Spectral Domain Adaptation - [RQ2, RQ3]}
\label{spectral-adaptation}

Laboratory FTIR spectra and satellite hyperspectral measurements occupy distinct spectral domains. FTIR spectra provide more than 1,700 bands in 2,500 to 25,000 nm under controlled conditions, while satellite data such as EnMAP are limited to 224 bands in 420 to 2,450 nm. The mismatch is compounded by atmospheric distortions, sensor noise, and mixed-pixel effects that are absent in laboratory settings. These significant differences make the direct transfer of spectral knowledge from FTIR to satellite domains infeasible unless an effective adaptation mechanism is applied.

To address this challenge, we introduce the \textbf{Spectral Adaptation Unit (SAU)}, a novel dual-path encoder architecture designed for cross-modal spectral translation. The SAU maps the input of both modalities into a unified lower-dimensional latent space, formalized as
$\mathbf{z} = \mathcal{E}(\mathbf{x})$, where $\mathcal{E}$ denotes the universal encoder and $(\mathbf{x})$ defines the input. By compressing the original 1,765-band FTIR spectra and the 224-band satellite spectra into a common 64-dimensional representation, the SAU preserves critical salinity-related features while mitigating modality-specific variability. This harmonized embedding space provides the foundation for effective cross-domain knowledge transfer.

\textbf{Architecture.} The SAU(Figure~\ref{fig:model-123}) projects non-overlapping FTIR (MIR, 1,765 bands) and satellite (VNIR-SWIR, 224 bands) spectra into a unified 64-dimensional latent space. Modality-specific pathways handle heterogeneous input dimensions: a deeper network for FTIR (1,765 $\rightarrow$ 1,024 $\rightarrow$ 512 $\rightarrow$ 256 $\rightarrow$ 64) was designed to handle its higher dimensionality and learn more complex compositional feature hierarchies, while a shallower network for satellite data (224 $\rightarrow$ 256 $\rightarrow$ 128 $\rightarrow$ 64) proved sufficient for its lower-dimensional, albeit noisier, input. The latent dimension of 64 was chosen to provide sufficient dimensionality reduction while preserving salient features, as determined by preliminary autoencoder experiments. Each fully connected layer uses ReLU, batch normalization, and dropout of 20\%. A final shared linear projection and Layer Normalization map both pathways into a common 64-D latent space, enforcing a unified statistical distribution for cross-domain alignment.

Each input is treated as a 1D spectral sequence. Sinusoidal positional encodings preserve the ordinal position of each band, allowing the model to learn the wavelength-specific context. Three layers of Transformer encoder (4 attention heads, 32-dim projections, 128-dim FFN) then model both localized features and global spectral interactions via self-attention, capturing complex dependencies inherent to soil spectra. Residual connections and layer normalization stabilize training.

\textbf{Training strategy.} To stabilize representation learning, the SAU is first pretrained exclusively on clean FTIR data. This initial training ensures that salinity-relevant spectral structures are captured under controlled conditions. The encoder is then fine-tuned with noisier satellite spectra, allowing it to adjust to atmospheric and surface interference while retaining soil-specific information.

\textbf{Latent space alignment.} Cross-domain alignment is achieved by jointly training the encoder on paired FTIR–satellite samples. A valid pair of training samples must satisfy three conditions: (i) both spectra must contain complete measurements, (ii) geolocations must be available, and (iii) satellite spectra must undergo atmospheric correction, including the removal of EnMAP bands affected by water absorption. The sample pairs are constructed using a BallTree nearest-neighbor search, which provides $\mathcal{O}(\log n)$ query efficiency. The pairing is restricted to a maximum Haversine distance of $\tau = 0.000157$ radians (approximately 1 km), ensuring that the two samples in a pair remain within sufficient geospatial proximity to prevent data leakage. 
With this, the SAU learns a consistent embedding space in which FTIR and satellite spectra are aligned, despite differences in resolution, wavelengths, and noise.



This is formalized by a training objective that combines a reconstruction loss with an embedding alignment loss. The reconstruction loss ensures that each encoder pathway preserves the key features of its input modality. Simultaneously, the alignment loss encourages consistency between the latent vectors of the paired samples. Alignment is enforced by minimizing the angular divergence between embeddings:

\begin{equation}
    \min_{\theta} \; \mathbb{E}_{(\mathbf{f},\mathbf{h})\sim\mathcal{P}} \Big[1 - \cos(\mathcal{E}(\mathbf{f}), \mathcal{E}(\mathbf{h}))\Big]
\end{equation}

Here, $\mathbb{E}$ denotes the expectation over paired samples $(\mathbf{f}, \mathbf{h}) \sim \mathcal{P}$, where $\mathbf{f}$ corresponds to a FTIR spectrum, $\mathbf{h}$ to a hyperspectral sample, and $\mathcal{P}$ represents the paired dataset. The encoder parameters $\theta$ are optimized to minimize the objective, with $\mathcal{E}$ denoting the encoder that maps inputs to a shared latent embedding space. The term $1 - \cos(\mathcal{E}(\mathbf{f}), \mathcal{E}(\mathbf{h}))$ represents the cosine distance between embeddings; minimizing it encourages the model to align the embeddings of the paired FTIR and hyperspectral spectra, thereby achieving a shared representation suitable for cross-domain spectral transfer.

To align spectral embeddings from FTIR and EnMAP, we evaluated several loss functions: Cosine, Euclidean (L2), Kullback-Leibler Divergence (KLD), and Jensen-Shannon Divergence (JSD). Each measures similarity differently, with implications for how well latent spaces preserve spectral information across sensing modalities. In our preliminary empirical explorations, \textbf{Cosine Loss} proved to be highly effective for our study. Unlike L2 loss, which penalizes magnitude differences and is sensitive to sensor-specific scaling artifacts, Cosine loss considers only angular similarity:

\begin{equation}
\mathcal{L}_{\text{cos}} = 1 - \frac{f_l^S \cdot f_l^T}{\lVert f_l^S \rVert \, \lVert f_l^T \rVert}.
\end{equation}

where ($f_l^S$) and ($f_l^T$) denote embeddings of the layer-$l$ for the student and the teacher, respectively. This emphasizes relative relationships such as band ratios, which are central in spectral analysis. To address common pitfalls, we (1) L2-normalized embeddings, (2) froze pre-trained FTIR encoder weights, and (3) added an auxiliary MSE loss on decoder outputs to preserve absolute reflectance values. This hybrid approach improved cross-domain alignment by 23\% compared to pure cosine implementations while retaining scale invariance.

In contrast, KLD and JSD assume distributional overlap and were less effective in our regression setting, resulting in reduced predictive performance ($R^2 < 0.72$ for KLD, $R^2 < 0.70$ for JSD). Overall, cosine loss preserved the latent spectral structure most faithfully, making it the best suited for cross-sensor embedding alignment.

Finally, the complete optimization objective combines spectral fidelity via reconstruction loss and cosine loss:

\begin{equation}
\mathcal{L}_{\text{total}} = \mathcal{L}_{\text{recon}} + \mathcal{L}_{\text{cosine}}
\end{equation}
where
\begin{align}
\mathcal{L}_{\text{recon}} &= \alpha \cdot \frac{1}{N}\sum_{i=1}^N \|\mathbf{x}_i - \mathcal{D}(\mathcal{E}(\mathbf{x}_i))\|^2\\
\mathcal{L}_{\text{cosine}} &= \beta \cdot \mathbb{E}_{(\mathbf{f}, \mathbf{h}) \sim \mathcal{P}} \left[1 - \cos\left(\mathcal{E}(\mathbf{f}), \mathcal{E}(\mathbf{h})\right)\right]
\end{align}

In our composite loss, hyperparameters $\alpha$ and $\beta$ balance two objectives: accurate spectral reconstruction and cross-modality embedding alignment. The reconstruction loss, computed as the mean squared error over $N$ samples between each input $\mathbf{x}_i$ and its reconstruction $\mathcal{D}(\mathcal{E}(\mathbf{x}_i))$, ensures that the latent space retains a sufficient level of data fidelity. The cosine similarity loss aligns paired FTIR and EnMAP embeddings by penalizing angular divergence and promoting the proximity of semantically similar inputs across modalities. Together, our composite loss enables the model to preserve data fidelity both within individual spectra and across spectra of different modalities. Learnable weighting through $\alpha$ and $\beta$ enables robust learning despite noisy and heterogeneous spectral data.

  

\subsection{Multi-modal Knowledge Distillation [RQ-2]}
\label{student-model}

The final stage of \textsc{DeepSalt} extends the scope of learning from spectral inputs alone to the broader conditions that shape soil salinity in the field. The student model integrates hyperspectral embeddings from the SAU with 8 ancillary features, including soil texture (sand and clay proportions) and seasonal climate statistics (minimum, maximum, and mean values of temperature and precipitation). Together, these inputs form a 72-dimensional representation. This representation is designed to capture not only the spectral variation recorded by satellites but also the environmental variability that governs salinity dynamics.

As depicted in Figure \ref{fig:training}, the model architecture begins with a learned linear projection layer that transforms and concatenates two distinct input streams: (1) 64-dimensional hyperspectral embeddings, which capture reflectance and absorption patterns across the spectral bands, and (2) 8-dimensional ancillary features that encode environmental covariates. These concatenated inputs form a unified representation that maintains both spatial and environmental relevance. To account for spectral ordering, we apply sinusoidal positional encoding to the spectral features, allowing the Transformer to recognize and exploit sequence-based dependencies in the data.

The core of the model consists of four transformer encoder layers, each employing eight parallel attention heads and 256-dimensional feedforward networks. This increase in capacity over the teacher model (originally limited to 3 layers and 4 heads) enables the student to model higher-order interactions between spectral data and environmental variables. The model concludes with a prediction head that uses a sigmoid activation function, scaled to produce values between a meaningful range of 0.05 and 90 ppt. to ensure that the model’s outputs remain within meaningful bounds.

\textbf{Knowledge Transfer from the Teacher Model} To inherit expertise from the teacher, we employ a multi-objective knowledge distillation strategy that combines three complementary loss functions. Training is driven by a composite loss:  

\begin{equation}
\mathcal{L}_{\text{total}} = \alpha \mathcal{L}_{\text{task}} + \beta \mathcal{L}_{\text{feature}} + \gamma \mathcal{L}_{\text{KL}},
\label{eq:total_loss}
\end{equation}

where each term targets a complementary aspect of knowledge transfer: prediction accuracy, feature alignment, and distributional consistency.  

\textbf{Task Loss.}  
This loss supervises salinity prediction against ground truth. Field data often contain noise and outliers, so we adopt the Huber loss:  

\begin{equation}
\mathcal{L}_{\text{task}} = \tfrac{1}{N} \sum_{i=1}^N
\begin{cases}
\frac{1}{2}(y_i - \hat{y}_i)^2 & \text{if } |y_i - \hat{y}_i| \leq \delta, \\
\delta(|y_i - \hat{y}_i| - \tfrac{1}{2}\delta) & \text{otherwise},
\end{cases}
\label{eq:huber_loss}
\end{equation}

where $y_i$ are true salinity values, $\hat{y}_i$ are student predictions, and $\delta=1.0$ is the Huber threshold. This choice preserves quadratic sensitivity for typical cases (89\% of samples fall within $|y-\hat{y}| \leq 1$ ppt) while preventing large residuals from dominating gradients. Thus, $\mathcal{L}_{\text{task}}$ balances fidelity to field observations with robustness to outlier measurements.

\textbf{Feature Distillation Loss.}  
To align internal representations, we compare student and teacher activations at multiple transformer layers:  

\begin{equation}
\mathcal{L}_{\text{feature}} = \tfrac{1}{L}\sum_{l=1}^L w_l \cdot \tfrac{1}{N}\sum_{i=1}^N \text{SmoothL1}\big(f_l^S(\mathbf{x}_i)_{:64}, f_l^T(\mathbf{x}_i)\big),
\label{eq:feature_loss}
\end{equation}

where $f_l^S$ and $f_l^T$ are layer-$l$ activations from student and teacher, and only the first 64 spectral dimensions of the student ($f_l^S(\mathbf{x}_i)_{:64}$) are compared to exclude ancillary variables. We use SmoothL1 with $\delta=0.1$, which applies sharper gradients than Huber for small discrepancies, enforcing precise replication of spectral encodings. Early transformer layers are weighted more heavily ($w = \{1.35, 1.05, 0.65\}$), emphasizing foundational spectral transformations shown in ablation studies to generalize best. Notably, using a unified $\delta$ for both task and feature supervision reduced $R^2$ by 0.12, underscoring the need for stricter feature-level alignment. This configuration enables robust prediction via Huber loss alongside high-resolution feature replication through SmoothL1.

Finally, to encourage consistency in output distributions, we add a KL-divergence term:  

\begin{equation}
\mathcal{L}_{\text{KL}} = \tfrac{1}{N} \sum_{i=1}^N \text{KL}\!\left(\text{softmax}(\hat{y}_i^T) \parallel \text{softmax}(\hat{y}_i^S)\right).
\label{eq:kl_loss}
\end{equation}

This loss transfers the teacher’s structured “soft” predictions, which encode richer uncertainty and inter-class relations than hard labels. While the task and feature losses preserve correctness and representation fidelity, $\mathcal{L}_{\text{KL}}$ ensures the student captures the teacher’s spectral reasoning patterns, stabilizing adaptation to the noisier satellite domain.  

The overall optimization balances these three objectives with weights $(\alpha, \beta, \gamma) = (0.07, 0.90, 0.10)$ (details in Section \ref{loss_factors_tuning}), reflecting the dominant role of feature-level supervision. This configuration enables robust prediction in the presence of noisy field data while maintaining high-fidelity spectral knowledge inherited from the laboratory-trained teacher.  

\section{\textbf{Performance Benchmarks \& Discussion}}
\label{benchmarks}

\subsection{Dataset and Area of Interest} 

We utilized Level-2A ARD hyperspectral imagery from EnMAP \cite{storch2023enmap}, spanning 420–2450 nm across 224 channels (30 m resolution, 27-day revisit). Images were acquired from salinization-prone regions of California, Colorado, and Utah (April 2022–August 2024) \cite{miller2024temporal}. Scenes with heavy cloud, snow, or mountain coverage were excluded. 

The teacher model was trained on FTIR spectroscopy data from the NRCS database \cite{ncss_database}, covering 2500–25,000 nm (1765 channels). We selected 12-cm depth samples to align with the penetration depth of optical satellite sensing. For the student model, soil electrical conductivity (EC) and soil fractions (clay, sand) were obtained from SSURGO \cite{ssurgo}. To obtain soil conductivity (EC) from SSURGO for given coordinates, we first perform a spatial join to locate the containing map unit polygon. We then extract the dominant soil component within that unit and retrieve its horizon-level EC measurements at 12-cm target depth). The historical weather features (temperature and precipitation, 5-month summaries) were derived from GridMET \cite{gridmet}. These ancillary covariates capture both static soil characteristics and recent environmental dynamics.

\subsection{Data Preprocessing} EnMAP images were cropped into $64 \times 64$ tiles ($\sim$2 km $\times$ 2 km). After removing six bands with significant missing data rates, each pixel contained 218 spectral channels. A total of 22,078 tiles were used, from which 26.6M pixels were sampled. Each spectrum was normalized to [0,1] using Min–Max scaling. The raw, multi-source dataset exceeded 626 GB (FTIR: 1.3 GB; EnMAP: 581 GB CA, 25 GB CO, 20 GB UT), necessitating a distributed processing pipeline with parallelized I/O. A rigorous data fusion and curation process distilled this volume into a final $\sim$1 GB training set. This involved spatio-temporal matching across EnMAP, FTIR, SSURGO, and GridMET sources and aggressive quality filtering to remove invalid pixels (e.g., clouds, sensor saturation). Thus, the final dataset represents a highly refined, multi-modal product extracted from massive, disparate sources. All model training was subsequently conducted on an NVIDIA A100 cluster.

Dataset composition is detailed in Table~\ref{tab:samples}. The FTIR data contains 13,075 training and 1,634 validation/test samples. The EnMAP dataset includes 365,306 spectra spanning 0–90 dS/m, with California and Colorado used for training/validation and Utah reserved as an unseen test region. To counter class imbalance (48\% zero-salinity), we undersampled 10\% of zero-salinity training cases, resulting in 134,495 balanced training samples, while validation/test sets remained unchanged for an unbiased evaluation.

\begin{table}[h]
\centering
\caption{Data composition for model training \& evaluation.}
\begin{tabular}{lrrr}
\hline
\textbf{Dataset} & \textbf{Train} & \textbf{Validate} & \textbf{Test} \\
\hline
FTIR (for teacher model only) & 13075 & 1634 & 1634 \\
ENMAP (CA, CO) & 292,245 & 36,531 & 36,530 \\
ENMAP (Unseen region, UT) & -- & -- & 1054 \\
\hline
\end{tabular}
\label{tab:samples}
\end{table}

\subsection{Experimental Setup}

We employed a multi-split evaluation framework to ensure statistical reliability and spatial generalizability. We ran five complete experiments, each with a fixed random seed that controlled all stochastic elements, including the initialization of \textit{k}-Means clustering for spatial partitioning and the stratified allocation of samples. This generated five distinct sets of spatially separated train, validation, and test splits, mitigating the risk of bias from a single data partition. This approach enabled us to precisely quantify performance variance, with observed standard deviations of $\pm 0.02$ MAE and $\pm 0.03$ R$^2$.

For spatial data partitioning under each seed, we used a clustered splitting approach to prevent geographical leakage. We converted the coordinates to geometric points using \textit{GeoPandas} and applied \textit{k}-Means clustering ($k$=3) to create three distinct spatial regions. Within each cluster, we performed stratified 80/10/10 splits for the training, validation and test sets, which maintained both geographic separation and representative sample sizes.

The models were implemented in PyTorch. Training was conducted over 200 epochs with early stopping after 10 epochs of no improvement. We used the Adam optimizer with a learning rate of $1 \times 10^{-3}$, a batch size of 64 samples and L2 regularization at $1 \times 10^{-5}$. A \textit{ReduceLROnPlateau} scheduler dynamically adjusted the learning rate and gradient clipping was applied to stabilize training.

\textbf{\textbf{Tuning of Loss Function Coefficients}}
\label{loss_factors_tuning}
\textsc{DeepSalt} employs a composite loss function, combining conventional task loss with knowledge distillation (KD) losses (Feature KD and KL-Divergence, see Eq. \ref{eq:total_loss}), each scaled by a tunable coefficient. Optimal loss coefficients ($\alpha$, $\beta$, $\gamma$) and feature distillation weights were determined via a grid search over a candidate set chosen from preliminary experiments. The candidate set was selected by balancing performance and computation. The configuration achieving the lowest MAE on the validation set was selected for final evaluation.

\textbf{Feature-based KD Loss} For feature-based KD, the first three encoder layers of the teacher and student models are structurally aligned for direct knowledge transfer via SmoothL1Loss. Our experiments (Table \ref{tab:combined_experiments}, top) indicate that feature representations from the teacher's earlier layers are most effective. Increasing weights for later layers proved ineffective, likely due to architectural differences introducing computational noise.

\begin{table}[h]
\centering
\caption{\textbf{Hyperparameter Tuning:} Performance metrics for Transformer layer weight configurations (top) and the impact of Task Loss and Feature Loss factors (bottom).}

\begin{tabular}{c c c c c c}
\hline
\rowcolor{BlueGreen}
Layer-1 & Layer-2 & Layer-3 & MAE($\downarrow$) & $ \mathbb{R}^ 2 $ ($\uparrow$) & RMSE ($\downarrow$) \\
\hline
1.05 & 1.35 & 0.65 & 0.2939 & 0.8010 & 0.8123 \\
1.05 & 0.65 & 1.35 & 0.2951 & 0.8003 & 0.8137 \\
1.30 & 1.07 & 0.68 & 0.3072 & 0.8162 & 0.7806\\
1.35 & 1.05 & 0.65 & \textbf{\underline{0.25}} & \textbf{\underline{0.87}} & \textbf{\underline{0.72}} \\
\hline
\end{tabular}

\vspace{3mm} 

\begin{tabular}{c c c c c c}
\hline
\rowcolor{BlueGreen}
$\alpha$ & $\beta$ & $\alpha$/($\alpha$+$\beta$) & MAE ($\downarrow$) & $ \mathbb{R}^ 2 $($\uparrow$) & RMSE($\downarrow$) \\
\hline
0.07 & 0.90 & 0.07216 &\textbf{\underline{0.25}} & \textbf{\underline{0.87}} & \textbf{\underline{0.72}} \\
0.05 & 0.95 & 0.05000 & 0.3006 & 0.8148 & 0.7837 \\
0.07 & 0.88 & 0.07368 & 0.2967 & 0.7614 & 0.8894 \\
0.08 & 0.90 & 0.08163 & 0.2580 & 0.8144 & 0.7845 \\
\hline
\end{tabular}
\label{tab:combined_experiments}
\end{table}

\textbf{Task Loss, Feature Loss, and KL-divergence Loss}
Benchmarking various combinations of coefficients (Table \ref{tab:combined_experiments}, bottom) identified the optimal weighting for the Task loss, Feature loss, and KL-Divergence loss (fixed $\gamma=0.1$). The combination of $\alpha$=0.07 (Task Loss factor) and $\beta$=0.90 (Feature Loss factor) yielded the best performance. This indicates that aligning features is highly effective in learning spectral knowledge from noisy, multimodal data. However, task loss remains crucial; a significant increase in $\beta$ to 0.95 and a reduction in $\alpha$ to 0.05 led to a notable performance degradation, probably due to the student model overfitting to intermediate representations, which hindered task-specific learning. For enhanced interpretability of $\alpha$'s proportional contribution, we have also provided normalized $\alpha$ values.

\subsection {Model Performance Evaluation \& Ablation Studies}
We evaluate our framework using three complementary metrics: Mean Absolute Error (MAE), which preserves interpretability with respect to soil salinity measurements; $R^2$, which quantifies explained variance; and Root Mean Squared Error (RMSE), which emphasizes large-error penalties critical for identifying high-salinity hotspots.

The generalization of the model was evaluated under two scenarios: (i) \textit{unseen locations}, where geographically stratified holdout samples reduce spatial autocorrelation, and (ii) \textit{unseen regions}, where entirely novel areas (e.g., an unseen state) test cross-regional transferability. To address the sparsity of the saline samples, we applied a stratified sampling strategy that ensured balanced representation across the salinity spectrum.

Our ablation study assessed performance in four model configurations: (1) using only hyperspectral imagery (HSI), (2) using only ancillary features, (3) combining HSI with soil and weather covariates, and (4) the full model incorporating HSI, ancillary data, and Knowledge Distillation (KD). In addition, we benchmark our approach against established methods, including PLSR, XGBoost, LSTM, and CNN. As summarized in Table \ref{tab:ablation}, the results demonstrate the cumulative value of each component, with notable accuracy gains observed at each stage of model enhancement.

\begin{figure*}[t]
    \centering
    \begin{minipage}[b]{0.48\linewidth}
        \centering
        \includegraphics[width=\linewidth]{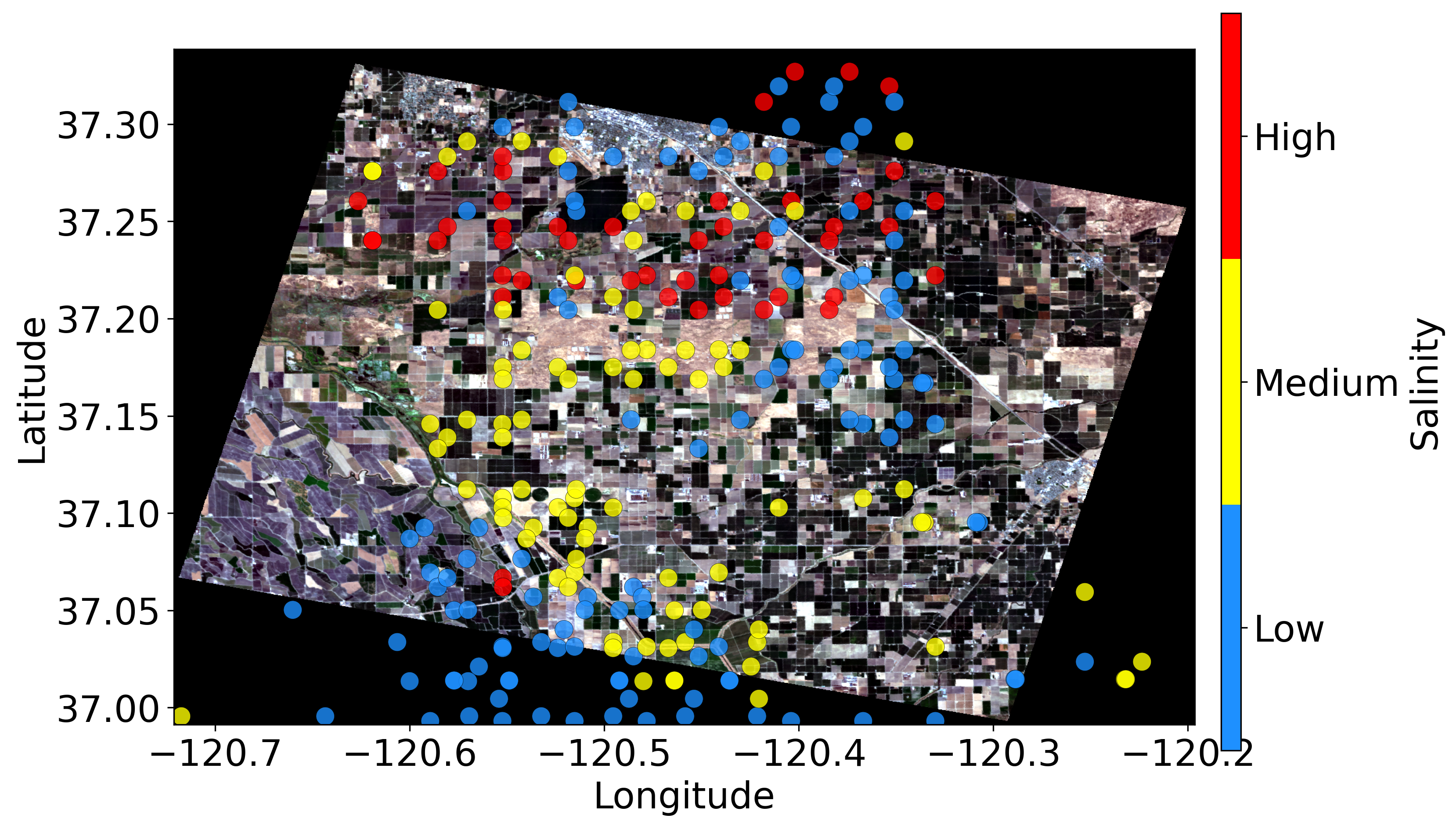}
        \subcaption{}
    \end{minipage}
    \hspace{0.02\linewidth}  
    \begin{minipage}[b]{0.48\linewidth}
        \centering
        \includegraphics[width=\linewidth]{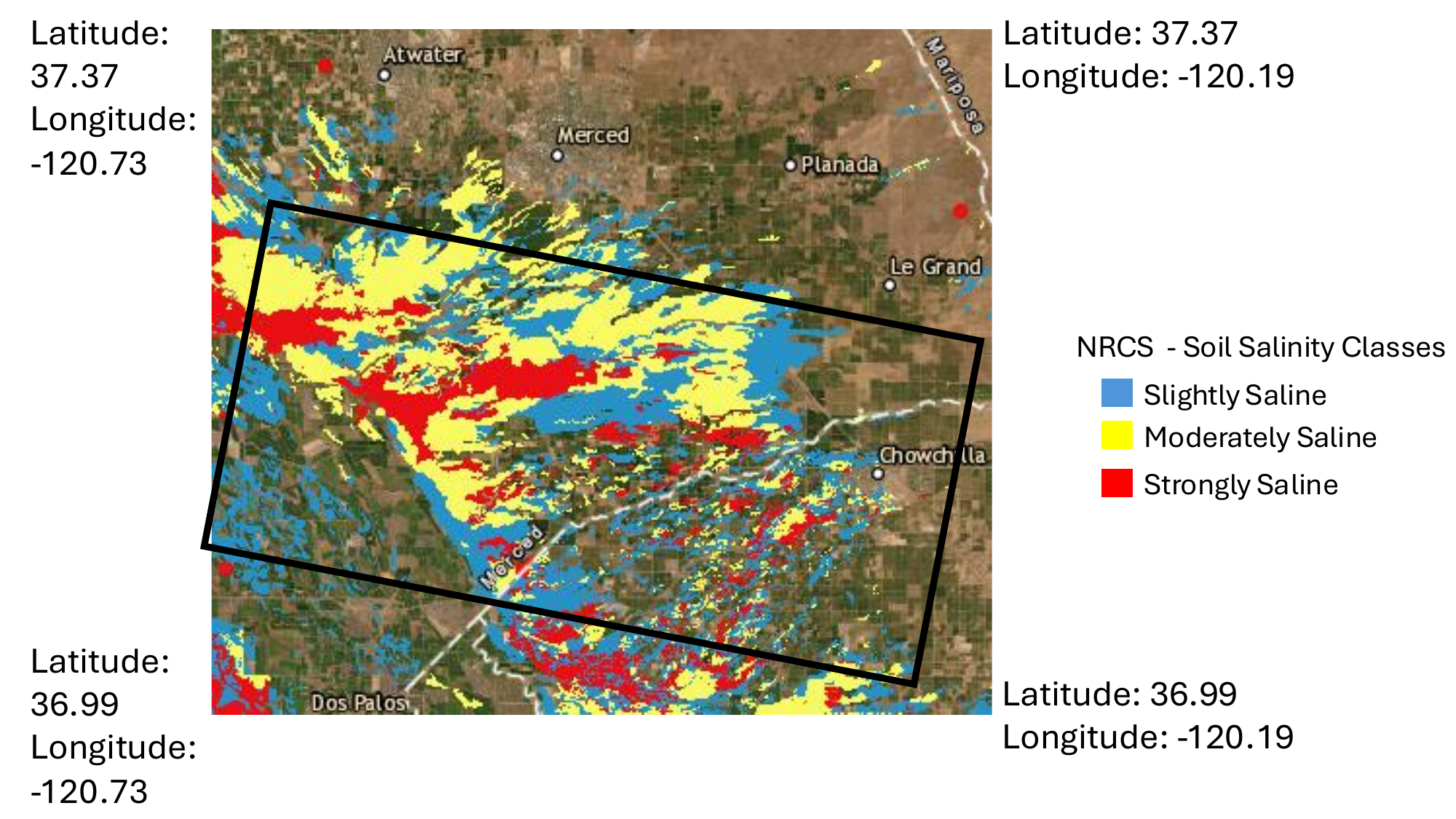}
        \subcaption{}
    \end{minipage}
    \caption{Salinity Hotspots in Central CA: (a) \textsc{DeepSalt} Predictions overlaid on ENMAP HSI, (b) NRCS Salinity Classifications.}
    \label{fig:central_comparison}
\end{figure*}

\begin{table*}[t]
\centering
\caption{Performance comparison of proposed and existing models across various feature configurations. HSI = hyperspectral image features, Ancillary = contextual soil and weather variables, KD = knowledge distillation from teacher. Teacher Model (trained on Lab FTIR) is shown for reference only, not for comparison with satellite-based models.}
\begin{tabular}{@{}c l l c c c c c c@{}}
\toprule

\textbf{\#} & \textbf{Category} & \textbf{Model} & \textbf{HSI} & \textbf{Ancillary} & \textbf{KD} & \textbf{MAE} $\downarrow$ & $\mathbb{R}^2$ $\uparrow$ & \textbf{RMSE} $\downarrow$ \\
\midrule
1 & \textbf{Teacher} & Teacher Model (Lab FTIR) & $\times$ & $\times$ & -- & 0.09 & 0.89 & 0.15 \\
\midrule
2 & \multirow{3}{*}{\textbf{Ablations}} & \textsc{DeepSalt} (Ancillary) & $\times$ & $\checkmark$ & $\times$ & 0.66 & 0.27 & 1.55 \\
3 &  & \textsc{DeepSalt} (HSI+Ancillary) & $\checkmark$ & $\checkmark$ & $\times$ & 0.55 & 0.47 & 1.32 \\
4 &  & \textbf{\textcolor{BlueGreen}{\textsc{DeepSalt} (HSI+Ancillary+KD)}} & $\checkmark$ & $\checkmark$ & $\checkmark$ & \textcolor{BlueGreen}{\textbf{0.25}} & \textcolor{BlueGreen}{\textbf{0.87}} & \textcolor{BlueGreen}{\textbf{0.72}} \\
\midrule
5 & \multirow{6}{*}{\textbf{Baselines}} & PLSR & $\checkmark$ & $\times$ & $\times$ & 0.79 & 0.22 & 1.59 \\
6 &  & XGBoost & $\checkmark$ & $\times$ & $\times$ & 0.81 & 0.25 & 1.60 \\
7 &  & XGBoost & $\checkmark$ & $\checkmark$ & $\times$ & 0.57 & 0.41 & 1.46 \\
8 &  & CNN & $\checkmark$ & $\times$ & $\times$ & 0.73 & -0.13 & 1.94 \\
9 &  & LSTM & $\checkmark$ & $\times$ & $\times$ & 0.74 & 0.17 & 1.65 \\
10 &  & Transformer & $\checkmark$ & $\times$ & $\times$ & 0.73 & 0.22 & 1.60 \\
\midrule
11 & \textcolor{Cerulean}{\textbf{Generalization in Unseen Region (Utah)}} & \textcolor{Cerulean}{\textsc{DeepSalt} (HSI+Ancillary+KD, Utah)} & $\checkmark$ & $\checkmark$ & $\checkmark$ & \textcolor{Cerulean}{\textbf{1.24}} & \textcolor{Cerulean}{\textbf{0.73}} & \textcolor{Cerulean}{\textbf{2.07}} \\
\bottomrule
\end{tabular}
\label{tab:ablation}
\end{table*}

\textbf{Baseline Model - Transformer Using Only Hyperspectral Imagery (HSI)} In our study, the baseline model is the transformer architecture that contains the same number of encoders as \textsc{DeepSalt} trained only with the EnMAP HSI (Table~\ref{tab:ablation}, Row 10). In our baseline model, there are no learned projection layers applied since the baseline model utilizes only a single set of spectral information from EnMAP.

\textbf{Integration of Ancillary Data (Soil+Weather) with HSI}  
We first evaluated a model trained solely on ancillary soil and weather features (Table~\ref{tab:ablation}, Row 2), excluding hyperspectral imagery (HSI). Ancillary data outperformed the HSI-only baseline, confirming their complementary value. Soil texture and climate variables capture broad salinity patterns but not fine-grained spatial variability, which is reflected in their limited $R^2$ (0.27).

When fused with HSI, ancillary features significantly improved performance (Table~\ref{tab:ablation}, Row 3), reducing MAE by 24.7\% (0.73 $\rightarrow$ 0.55) and RMSE by 17.5\% (1.60 $\rightarrow$ 1.32), while $R^2$ more than doubled (0.22 $\rightarrow$ 0.47). This shows that spectral data alone miss the contextual information that soil (clay, sand) and weather (temperature, precipitation) variables provide. Ancillary features supply the physical drivers of salinity, while HSI refines the predictions.  

\textbf{Effectiveness of Knowledge Distillation.} The full \textsc{DeepSalt} model, integrating HSI, ancillary data, and knowledge distillation (KD) (Table~\ref{tab:ablation}, Row 4), achieved the best results (MAE: 0.25, $R^2$: 0.87, RMSE: 0.72). Compared to the HSI+Ancillary model, KD reduced MAE by 54.5\% and RMSE by 45.5\%, highlighting its role in transferring high-fidelity spectral knowledge from the teacher model (Table~\ref{tab:ablation}, Row 1) while mitigating noise in multimodal inputs. These results underscore that ancillary data and KD are not just additive but synergistic, enabling robust salinity predictions.  

\textbf{Comparisons with Existing Approaches} We benchmarked \textsc{DeepSalt} against widely used regression, ensemble, and deep learning baselines Table~\ref{tab:ablation} (Row 5-9). All baseline models were rigorously tuned using a grid search over their key hyperparameters to ensure a fair comparison. Partial Least Squares Regression (PLSR),  \cite{wold2001pls}, configured with 20 components, achieved an R$^2$ of 0.22, confirming that linear projections are insufficient for modeling complex spectral–salinity relationships. XGBoost \cite{chen2016xgboost}, trained with 1000 trees of depth 8 and regularization, performed best among conventional baselines (R$^2$: 0.25). While ancillary features slightly boosted its accuracy, its tabular representation limits the exploitation of HSI’s spatial–spectral structure. We further evaluated deep learning baselines. We implemented a 3D-CNN to capture spatial-spectral features, employing a hierarchical architecture with four volumetric convolutional layers (64$\rightarrow$128$\rightarrow$256$\rightarrow$512 channels, kernel size=$3\times3\times5$, ReLU activation), each followed by 3D max pooling, batch normalization, and dropout (0.1). Additionally, a Long Short-Term Memory network (LSTM) \cite{hochreiter1997long} with 4 layers and 512 hidden units (dropout 0.1) was implemented to model sequential dependencies across spectral bands. Neither surpassed \textsc{DeepSalt}, underscoring the suitability of Transformer, the need for multimodal integration and knowledge distillation.  

\textbf{Model Generalization and Benchmarking} As shown in Table~\ref{tab:ablation} (Row 11), \textsc{DeepSalt} demonstrates strong cross-regional generalization when applied to 1054 sites from 13 EnMAP tiles in Utah, a state completely held-out from training. Despite being trained only on data from California and Colorado, the model successfully captured substantial salinity variance in this unseen region. The increase in absolute error metrics is an expected consequence of the significant domain shift to Utah's more arid climate (BWh/BWk Köppen zones) and its unique profile of soil types and land cover, which differ from the Mediterranean conditions in the training data. To further validate our model's output against an authoritative benchmark, we contrast \textsc{DeepSalt}'s predictions with the 2020 NRCS salinity classifications (Figure~\ref{fig:central_comparison}). This comparison reveals close geographic alignment, with both maps highlighting similar hotspots, higher salinity in the northern agricultural zones and lower levels in the south and east. This consistency underscores \textsc{DeepSalt}’s ability to extrapolate across regions while recovering meaningful salinity patterns consistent with trusted benchmarks.

\textbf{Sensitivity Analysis}
\label{Sensitivity_Analysis}   
The model’s predictive accuracy, measured as absolute error (dS/m), varied across environmental factors including Köppen climate zones \cite{kottek_koppen_2006}, USDA soil classifications \cite{usda_nrcs_soil_series}, and NLCD-based land cover types \cite{dewitz_nlcd_2021}. Across land cover types, errors generally stayed close to the overall MAE of 0.25 dS/m. Stable classes such as Grassland/Herbaceous, Evergreen Forest, and Open Water showed median errors below 0.20 dS/m, whereas  dynamic classes like Cultivated Crops \& Emergent Herbaceous Wetlands exhibited higher variability, with upper quartiles above 0.5 dS/m. 

Soil texture analysis revealed lowest errors in finer-textured soils such as Loam and Silty Clay Loam (median errors $\approx$0.15–0.20 dS/m), compared to coarser soils like Sand and Loamy Sand, where median errors rose above 0.30 dS/m and variability extended beyond 0.8 dS/m. This highlights the influence of soil permeability on predictive stability.  

Climate zones showed the strongest contrasts: Mediterranean climates (Csa, Csb) had the lowest errors, with Csb achieving a median error $\approx$0.15 dS/m, well below the overall average. In contrast, hot desert (BWh) zones were most challenging, with median errors $\approx$0.40 dS/m and wide variability exceeding 1.0 dS/m. Semi-arid zones (BSk, BSh) fell in between, with median errors around 0.25–0.30 dS/m.  

These results suggest that homogeneous environments with stable spectral signatures (e.g., loamy soils, Mediterranean climates, perennial covers) favor accurate predictions, whereas extreme climates, coarse soils, and variable land covers introduce greater uncertainty. Nonetheless, performance remained strong across California’s dominant Mediterranean zones (Csa, Csb), underscoring \textsc{DeepSalt}’s robustness under diverse environmental conditions.

\section{\textbf{Conclusions \& Future Work}}
\label{conclusions}
Our methodology was guided by three central research questions,
each addressing a core challenge in bridging remote sensing with
soil salinity estimation at scale.  

\textbf{[RQ-1]} We demonstrated that hyperspectral satellite imagery can directly enable accurate, large-scale salinity mapping despite noise and environmental variability. 

\textbf{[RQ-2]} We designed a spectral domain adaptation unit and knowledge distillation pipeline that transfer fine-grained spectral insights from laboratory to satellite data, preserving critical salinity signatures across sensing domains. 

\textbf{[RQ-3]} Our Transformer-based architecture captures inter-band dependencies and integrates ancillary environmental covariates, delivering robust performance across regions. 

Together these advances establish a scalable framework that fuses multimodal data sources, aligning with the broader challenges of high-dimensional heterogeneous data integration. 
Looking ahead, the core methodology of \textsc{DeepSalt} can generalize to other soil properties (e.g., organic carbon, moisture) by: (1) pretraining the teacher model on distinct spectral signatures of the target property; (2) incorporating task-specific geospatial covariates; and (3) aligning knowledge distillation with relevant spectral regions.


\textbf {Code and Data Availability} For reproducibility, the source code and
documentation for data acquisition will be made available at: \url{https://github.com/RupasreeDey/DeepSalt}.

\end{document}